\documentclass{article}
\usepackage{spconf,amsmath,epsfig}
\usepackage{paralist}
\usepackage{url}
\usepackage{color}

\begin{document}\sloppy

\def\x{{\mathbf x}}
\def\L{{\cal L}}

\title{Visual Summary of Egocentric Photostreams \\by Representative Keyframes}
%
\twoauthors
 {Marc Bola\~{n}os, Estefan\'ia Talavera and Petia Radeva}
	{Universitat de Barcelona\\
  \{marc.bolanos,etalavera,petia.radeva\}@ub.edu}
{Ricard Mestre and Xavier Gir\'o-i-Nieto\sthanks{This work has been developed in the framework of the project BigGraph TEC2013-43935-R, funded by the Spanish Ministerio de Economía y Competitividad and the European Regional Development Fund (ERDF). We gratefully acknowledge the support of NVIDIA Corporation with the donation of the GeoForce GTX Titan Z used in this work.}}  
  	{Universitat Politecnica de Catalunya\\
	xavier.giro@upc.edu}

\maketitle

\begin{abstract}
Building a visual summary from an egocentric photostream captured by a lifelogging wearable camera is of high interest for different applications (e.g. memory reinforcement). 
In this paper, we propose a new summarization method based on keyframes selection that 
uses visual features extracted by means of a convolutional neural network. Our method applies an unsupervised clustering for dividing the photostreams into events, and finally extracts the most relevant keyframe for each event. We assess the results by applying a blind-taste test on a group of 20 people who assessed the quality of the summaries.

\end{abstract}

\section{Introduction}
\label{sec:intro}

Lifelogging devices offer the possibility to record a rich set of data about the daily life of a person. A good example of this are wearable cameras, that are able to capture images from an egocentric point of view, continuously and during long periods of time. The acquired set of images comes in two formats depending on the device used: 1) high-temporal resolution videos, which usually produce more than 30fps and capture a lot of dynamical information, but they are only capable of storing some hours of data, or 2) low-temporal resolution photostreams, which usually produce only 1 or 2 fpm, but are able to capture events that happen during a whole day (having around 16 hours of autonomy).

Being able to automatically analyze and understand the large amount of visual information provided by these devices would be very useful for developing a wide range of applications. Some examples could be building a nutrition diary based on what, where and in which conditions the user eats for keeping track of any possible unhealthy habit, or providing an automatic summary of the whole day of the user for offering a memory aid to mild cognitive impairment (MCI) patients by reactivating their memory capabilities \cite{hodges2006sensecam}.

\begin{figure}[!ht]
  \centering
  \includegraphics[width=1\columnwidth]{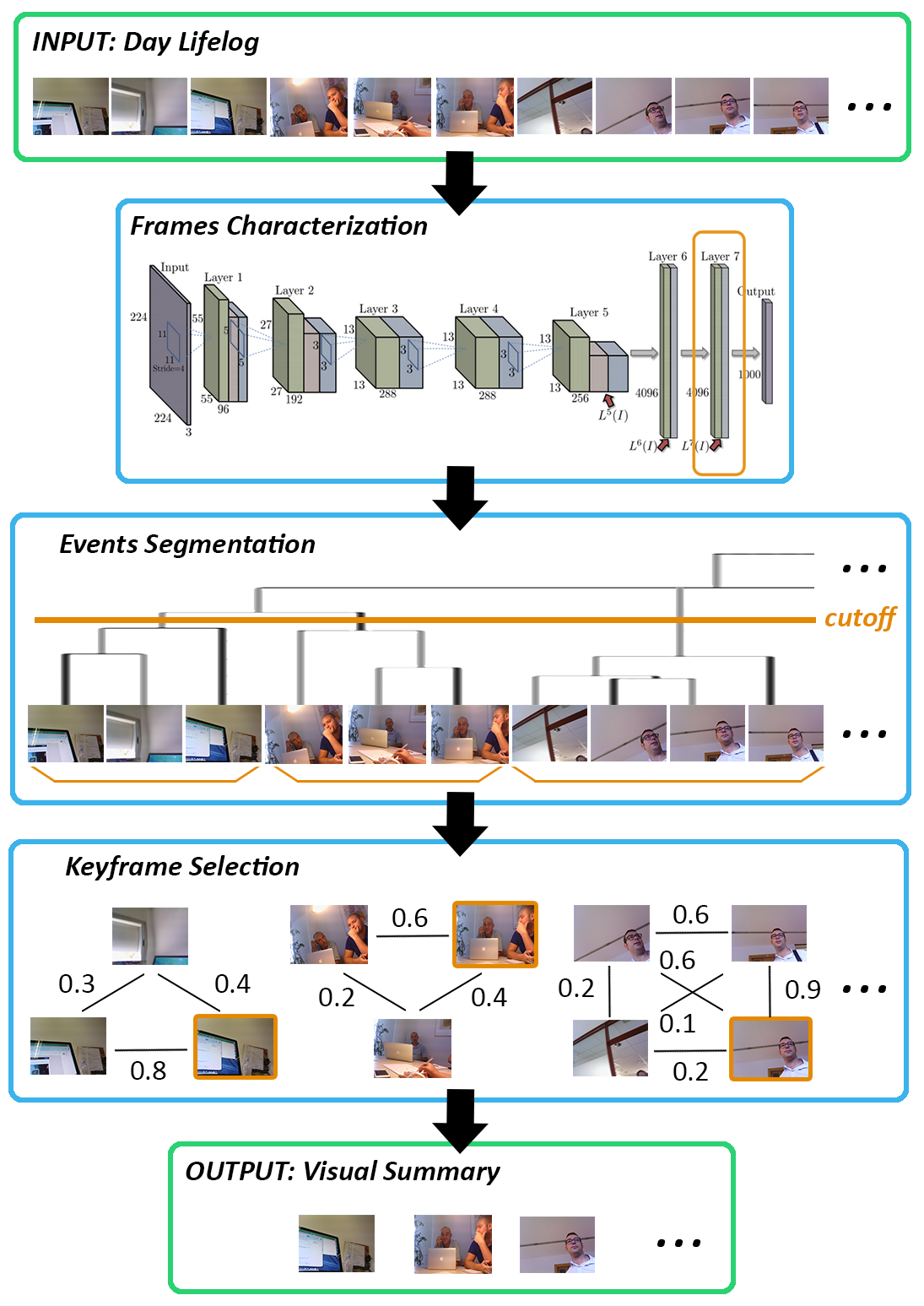}
\caption{Scheme of the proposed visual summarization.}
\label{fig:scheme_methodology}
\vspace{-1em}
\end{figure}


In our work, we focus on automatic extraction of a good summary that can be used as a memory aid for MCI patients. Usually, these patients suffer neuron degradation that generates them problems to recognize  familiar people, objects and places
\cite{lee2007providing}. Hence, the visual summary, automatically extracted, should be clear and informative enough to recall the daily activity with a simple glimpse.




In order to take into account our ultimate goal, we propose an approach that starts by extracting a set of features for frames characterization by means of a convolutional neural network.
These visual descriptors are used to segment events by running an agglomerative clustering, which is post-processed to guarantee a temporal coherency (similar to \cite{lin2006structuring}). 
Finally, a representative keyframe for each event is selected using the Random Walk  \cite{pearson1905problem} or Minimum Distance \cite{doherty2008investigating} algorithms. The overall scheme is depicted in Figure \ref{fig:scheme_methodology}.

This paper is structured as follows. Section \ref{sec:RelatedWork} overviews previous work for event segmentation and summarization in the field of egocentric video. Our approach is described in Section \ref{sec:method} and its quantitative and qualitative evaluation in Section \ref{sec:results}. Finally, Section \ref{sec:conclusions} draws the final conclusions and outlines our future work.
\section{Related Work}
\label{sec:RelatedWork}

The two main problems addressed in this paper, event segmentation and summarization, have been addressed in related egocentric data works, as presented in this section. 

\subsection{Egocentric event segmentation}

Most existing techniques agree that the first step for a summary construction is a shot- or event-based segmentation of the photostream or video. 
Lu and Grauman in \cite{lu2013story} and Bola\~nos et al. in \cite{bolanos2014video} both propose event segmentation that relies on motion information, colour and blurriness, integrated in an energy-minimization technique. The result is final event segmentation that is able to capture the different motion-related events that the user experiences. In the former approach \cite{lu2013story}, the authors use high-temporal videos and an optical flow descriptor for characterizing the neighbouring frames. In the latter one \cite{bolanos2014video}, instead of working with low-temporal data, a SIFT-flow descriptor is used, as it is more robust for capturing long-term motion relationships.
Poleg et al. \cite{poleg2014temporal} also propose motion-based segmentation, but they use a new method of Cumulative Displacement Curves for describing the motion between neighbouring video frames. The proposed solution is able to focus on the forward user movement and removes the noise of the head motion produced by head-mounted wearable cameras.
Other methods have been proposed using low-level sensor features like the work in \cite{Doherty2008automatically} that splits low-temporal resolution lifelogs in events. Lin and Hauptmann \cite{lin2006structuring} also propose a simple approach based on using colour features in a Time-Constrained K-Means clustering algorithm for keeping temporal coherence.
In \cite{talavera2015r-clustering}, Talavera et al. design a segmentation framework also based on an energy minimization framework. In this case, the authors offer the possibility to integrate different clustering and segmentation methods, offering more robust results. 

\subsection{Egocentric summarization}

Focusing on the summarization of lifelog data after event segmentation, there are two basic research directions, both of them aiming at removing those data, which are redundant or low-informative. In the case of video recordings, it is a common practice the select a subset of video segments to create a video summary. On the other hand, when working with devices that take single pictures at a low frame rate, the problem is usually tackled by selecting the most representative keyframes.
The most relevant work in the literature following the video approach is from Grauman et al. in \cite{lu2013story, lee2012discovering}, where a summary methodology for egocentric video sequences is proposed. The authors rely on an initial event segmentation, followed by the detection of salient objects and people, create  a graph linking events and the important objects/people, and finish with a selection of a subset of the events of interest. This final selection is based on combining three different measures: 1) \textit{Story} (choosing a set of shots that are able to follow the inherent story in the dataset), 2) \textit{Importance} (aimed at choosing only shots that show some important aspect of the day) and 3) \textit{Diversity} (adding a way to avoid repeating similar actions or events in the summary).
When considering the keyframe selection approach, one of the most relevant works is by Doherty et al. \cite{doherty2008investigating}, where the authors  study various selection methods like: 1) getting the frame in the middle of each segment, 2) getting the frame that is the most similar w.r.t. the rest of the frames in the event, or 3) selecting the closest frame to the event average.

\vspace{-1em}
\section{Methodology}
\label{sec:method}

This section presents our methodology for keyframe-based summarization of egocentric photostreams, depicted in Figure\;\ref{fig:scheme_methodology}. We start by characterizing each of the lifelog frames with a global scale visual descriptor. These features are used to create a visual-based event segmentation, which incorporates a post-processing step to guarantee time consistency. Finally, the most visually repetitive frame is selected as the most representative of the event.

\vspace{-1em}
\subsection{Frames characterization}

Convolutional Neural Networks (convnets or CNNs) have recently outperformed hand-crafted features in several computer vision tasks \cite{lecun1989backpropagation, NIPS2012_4824}. These networks have the ability to learn sets of features optimised for a pattern recognition problem described by a large amount of visual data.

The last layer of these convnets is typically a soft-max classifier, which in some works is ignored, and the penultimate fully connected layers are directly used as feature vectors. These visual features have been successfully used as any other traditional hand-crafted features for purposes such as image retrieval \cite{babenko2014neural} or classification \cite{Chatfield14devil}. 

In the field of egocentric video segmentation, convnets have also been proved as suitable for clustering purposes \cite{talavera2015r-clustering}. For this reason, we used a set of features extracted by means of the pre-trained \emph{CaffeNet} convnet included in the Caffe library \cite{jia2014caffe}. This convnet was inspired by \cite{NIPS2012_4824} and trained on ImageNet \cite{deng2009imagenet}. In our case, we used as features the output of the penultimate layer, a fully connected layer of 4,096 components, discarding this way the final soft-max layer, which was intended to classify 1,000 different semantic classes from ImageNet.

\begin{figure*}[!ht]
  \centering
  \includegraphics[width=1\textwidth]{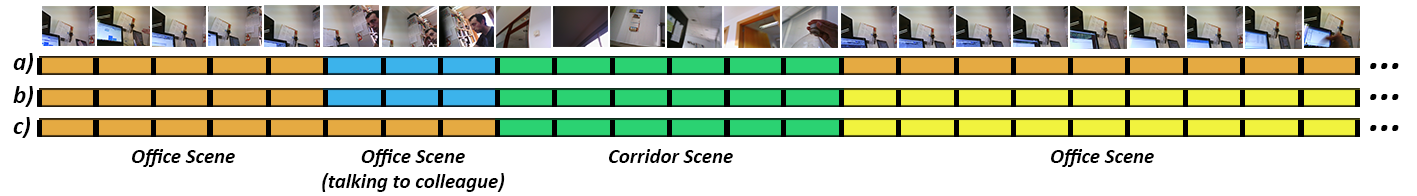}
\caption{Example of the events labeling produced by a) simply using the AC algorithm, b) applying the division strategy and c) additionally applying the fusion strategy. Each color represents a different event.}
\label{fig:division_fusion}
\end{figure*}

\subsection{Events segmentation}

The egocentric photostream is segmented with an unsupervised hierarchical Agglomerative Clustering (AC) \cite{day1984efficient} based on the convnet visual features. As proved in \cite{talavera2015r-clustering}, this clustering methodology reaches a reasonable accuracy for this task. In this way, we can define sets of images, each of them representing a different event. AC algorithms can be applied with different similarity measures. Different configurations were tested (see details in Section \ref{sec:results_clustering_evaluation}) and the best approach was obtained with the \textit{average} linkage method with Euclidean distance. This option determines the two most similar clusters to be fused in each iteration using the following distance:
\begin{eqnarray} \label{equ:AC}
\centering
arg\,min_{C_i,C_j \in {\bf C_t}} \ D(C_i,C_j), \mbox{ where} 
\end{eqnarray}
 \begin{eqnarray} \label{equ:average}
D(C_i,C_j) = \frac{1}{|C_i| \times |C_j|} \sum_{s_{k,i} \in C_i,s_{l,j} \in C_j}  \sqrt{f(s_{k,i})^2 - f(s_{l,j})^2},
\end{eqnarray}
where ${\bf C_t}$ is the set of clusters at iteration $t$, $s_{k,i}$ and $s_{l,j}$ are the samples in cluster $C_i$ and $C_j$, respectively, and $f(s)$ are the visual features extracted by means of the convnet.

However, creating the clusters based only on visual features often generates non-consistent solutions from a temporal perspective. Typically, images captured in the same scenario will be visually clustered as a single event despite corresponding to separate moments. For example, frames from the beginning of the day, (e.g. when the user takes the train for commuting to work) may be visually indistinguishable with other frames from the end of the day (e.g. when the user is going back home by train too). Additionally, another usual problem when relying only on visual features is that sometimes very small clusters can be generated, a result which should be avoided because an event is typically required to have a certain span in time (e.g. 3 minutes, in our work).

In order to solve these problems, we introduce two post-processing steps for refining the resulting clusters: \textit{Division} and \textit{Fusion}. The \textit{Division} step splits in different events those images in the same cluster which are temporally interrupted by events defined in other clusters. For example, the event in orange from Figure \ref{fig:division_fusion} a) is divided in two events (orange and yellow) in Figure \ref{fig:division_fusion} c) due to a \textit{Corridor scene} event (in green) interrupting the original \textit{Office scene}. On the other hand, the second post-processing step, \textit{Fusion}, will merge all those events shorter than a threshold with the closest neighboring event in time.


\subsection{Keyframe selection}

Once the photostream is split into the events, the next step is to carefully select a good subset of keyframes. To do so, we explored two different methods: \textit{random walk} and a \textit{minimum distance} approach. Both approaches are based on the assumption that the best photo to represent the event is the one, which is the most visually similar with the rest of the photos in the same cluster. As a result, each event can be automatically represented by a single image and, when all images combined, they will provide a visual summary of the user's day.

\subsubsection{Random Walk}

We propose to use the Random Walk algorithm \cite{pearson1905problem} in each of the events, separately. As a result, the algorithm will select the photo, which is more visually similar to the rest of the photos in the event. After applying the same procedure for all the events, we can have a good general representation of the main events that happened in the user's daily life.

The Random Walk algorithm works as follows: 1) the visual similarity for each pair of photos in the event is computed; 2) a graph described by a transitional probabilities matrix is built using the extracted similarities as weights on each of the edges; 3) the matrix eigenvectors are obtained, and 4) the image associated to the largest value in the first eigenvector is considered as the keyframe of the event.

\vspace{-1em}
\subsubsection{Minimum distance}

The second considered option selects the individual frame with the minimal accumulated distance with respect to all the other images in the same event. That is, let us consider the adjacency matrix $A=\{ a_{i,j}\}=\{ d_{s_i,s_j} \}$, where $d_{s_i,s_j}$ is the Euclidean distance between the descriptors of images $s_i$ and $s_j$ extracted by the convnet, $i=1,...N$, $j=1,...N$, where $N$ is the number of frames of the event. Let us consider the  vector $v=(\sum_j  a_{i,j})$ of accumulated distances. One can easily see that the index of the minimal component of vector $v$ i.e. $k=arg min_i \{v_i \}, \; i=1,...N$ 
determines the closest frame to the rest of frames in the corresponding event with respect to the $L_1$ norm \cite{doherty2008investigating}.

\section{Results}
\label{sec:results}

This section presents the quantitative and qualitative experiments run on a home-made egocentric dataset to assess the performance of the presented technique.

\subsection{Dataset}

Our experiments were performed on a home-made dataset of images acquired with a Narrative\footnote{\url{www.getnarrative.com}} wearable camera. This device is typically clipped on the users' clothes under the neck or around the chest area. The dataset, we used, is a subset of the one used by the authors in \cite{talavera2015r-clustering} (not using the SenseCam sets). It is composed of 5 day lifelogs of 3 different persons and has a total of 4,005 images. Furthermore, it includes the ground truth (GT) events segmentation for assessing the clustering results.

\subsection{Quantitative evaluation of event segmentation}
\label{sec:results_clustering_evaluation}

The first test assessed the quality of the photostream segmentation into events. In order to make this evaluation, we used the Jaccard Index, which is intended to measure the overlap of each of the resulting events and the GT the following way:
\begin{eqnarray} \label{equ:jaccard}
J(E, GT) = \sum_{e_i \in E \ g_j \in GT} M_{ij} \frac{e_i \cap g_j}{e_i \cup g_j},
\end{eqnarray}
where $E$ is the resulting set of events, $GT$ is the ground truth, $e_i$ and $g_j$ are a single event and a single GT segment respectively, and $M_{ij}$ is an indicator matrix with values $1$, iff $e_i$ has the highest match with $g_j$.

\begin{figure}[t]
  \centering
  	\includegraphics[width=0.8\columnwidth]{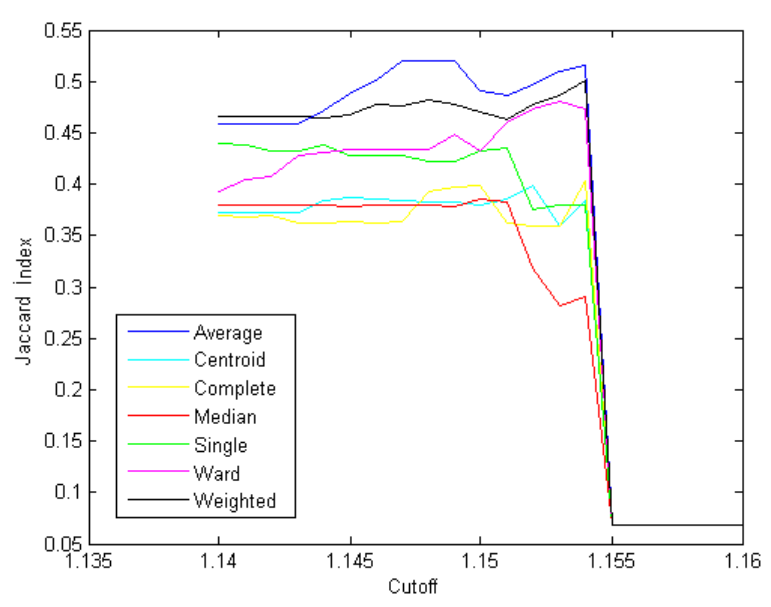}
	\caption{Average Jaccard index value obtained for the 5 sets. We compare each of the methods after applying the division-fusion strategy with respect to the best cut-off AC values.}
    \label{fig:AC_methods}
\end{figure}

\begin{figure}[t]
  \centering
  	\includegraphics[width=0.8\columnwidth]{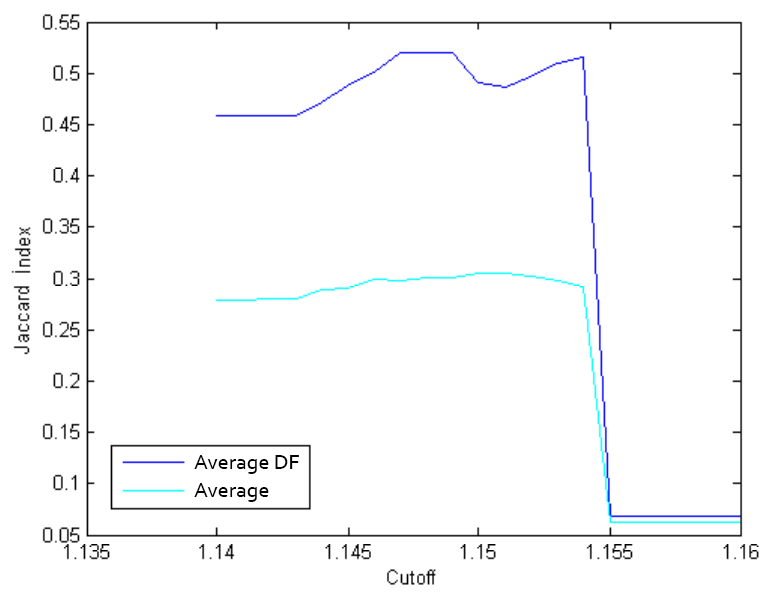}
	\caption{Effect when using (dark blue) the division-fusion (DF) strategy and when not using it (light blue) in the average Jaccard index result for all the sets.}
    \label{fig:AC_DF}
\end{figure}

We compared  different cluster distance methods with respect to the chosen cut-off parameter (which determines how many clusters are formed considering their distance value) for the AC (see Figure \ref{fig:AC_methods}). We choose the "average" with $\text{cutoff} = 1.154$ as the best option and, with this configuration, we  measured the gain of introducing the Division-Fusion strategy, illustrated in Figure \ref{fig:AC_DF}.


\subsection{Qualitative evaluation with blind-test taste}

\begin{figure}[t]
  \centering
  \includegraphics[width=0.8\columnwidth]{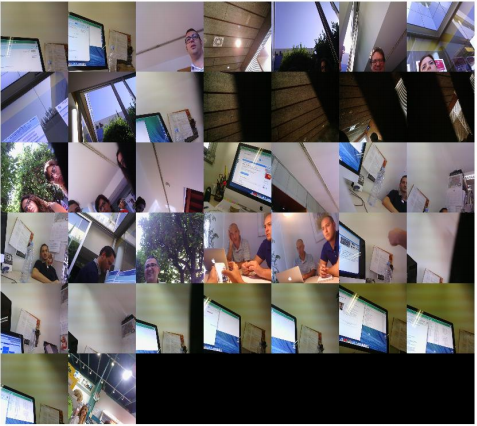}
\caption{Example of one of the summaries obtained by applying our approach on a dataset captured with Narrative camera.}
\label{fig:visual_summary_result}
\end{figure}

The assessment of visual summaries of a day, like the one shown in Figure \ref{fig:visual_summary_result}, is a challenging problem, because there is not a single solution for it. Different summaries of the same day may be considered equally satisfactory due to near duplicate images and subjectivity in the judgments. Therefore, we followed an evaluation procedure similar to the one adopted by Lu and Grauman \cite{lu2013story}. We designed a blind-taste test and asked to a group of 20 people to rate the output of different solutions, without knowing which of them corresponded to each configuration. 

\subsubsection{Keyframe selection}

The first qualitative evaluation focused on the keyframe selection strategy, comparing both presented algorithms (\textit{Random Walk} and \textit{Minimum Distance}) with a third one, \textit{Random Baseline}. In this first part, the three selection strategies were applied on each of the events defined by the GT annotation. 

\begin{figure}[t]
  \centering
  	\includegraphics[width=0.8\columnwidth]{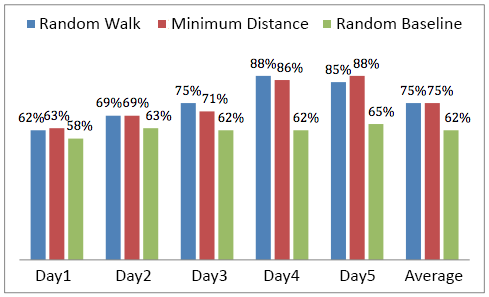}
	\caption{Results answering "yes" to the question "Is this image representative for the current event?"}
    \label{fig:results_question1}
\end{figure}

\begin{figure}[t]
  \centering
  	\includegraphics[width=0.8\columnwidth]{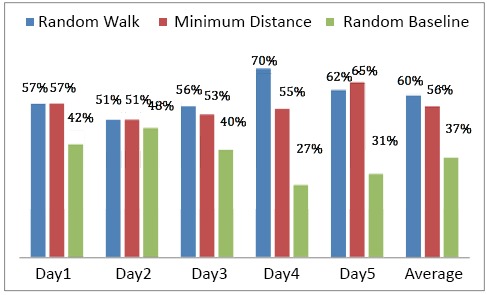}
	\caption{Results to the question "Which of the previous frames is the most representative for the event?"}
    \label{fig:results_question2}
\end{figure}

On the first part, we showed to the user a complete event according to the GT labels and, afterwards, the three keyframes selected by the three methods under comparison in a random sorting\footnote{If any of the results for the different methods was repeated, only one image was shown and the results were counted for both methods.}. Then, the user had to answer if each of the candidates was representative of the current event (results in Figure \ref{fig:results_question1}), and also choose which of them was the best one (results in Figure \ref{fig:results_question2}). This procedure was applied on each of the events of the day and results averaged per day.

Scoring results presented in Figures \ref{fig:results_question1} and \ref{fig:results_question2} indicate how both proposed solutions consistently outperform the random baseline for each day. The difference is more remarkable, when we asked the user to choose between only one of the possibilities (Figure \ref{fig:results_question2}). 
We must note that usually the result was very similar either for the Random Walk and the Minimum Distance, since in most of the cases both algorithms selected the same keyframe.


\subsubsection{Visual daily summary}

In the second part of our qualitative study, we assessed the whole daily summary, built with the automatic event segmentation and the different solutions for keyframe selection. In this experiment, we added a fourth configuration that built a visual summary with a temporal \textit{Uniform Sampling} of the day photostream, in such a way that the total amount of frames was the same as the amount of events detected through AC.

\begin{figure}[t]
  \centering
  	\includegraphics[width=0.8\columnwidth]{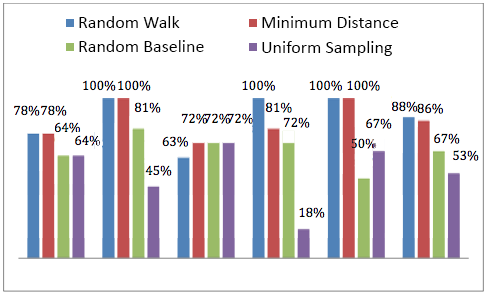}
	\caption{Results answering "yes" to the question "Can this set of images represent the day?"}
    \label{fig:results_question3}
\end{figure}

\begin{figure}[t]
  \centering
  	\includegraphics[width=0.8\columnwidth]{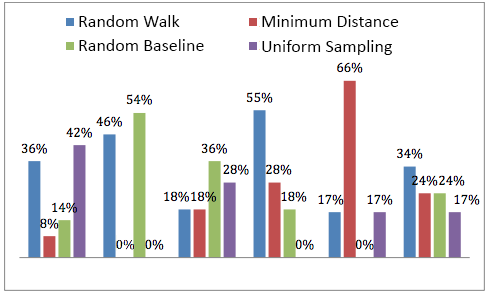}
	\caption{Results to the question "Which of the previous summaries does better describe the day?"}
    \label{fig:results_question4}
\end{figure}

This time the user was shown the four summaries of the day generated by the four configurations. Figure \ref{fig:visual_summary_result} provides an example built with the \textit{Random Walk} solution. For each summary, the user was firstly asked whether the set could represent the day (results in Figure \ref{fig:results_question3}), and also which of the four was the one that better described the day (results in Figure \ref{fig:results_question4}).

Focusing on the average results in Figure \ref{fig:visual_summary_result}, we can state that, either applying \textit{Random Walk} (88\%) or \textit{Minimal Distance} (86\%), most of the generated summaries were positively assessed by the graders. Moreover, when it comes to choose only the best summary, our method gathered 58\% of the total votes if we consider that the voting is exclusive and that the summaries produced by the Random Walk and the Minima Distance methods are very similar. As a result, we obtained 34\%  and 41\% of improvement respectively w.r.t the Random and the Uniform baselines.

\vspace{-1em}
\section{Conclusions}
\label{sec:conclusions}

In this work, we  presented a new methodology to extract a keyframe-based summary from egocentric photostreams. After the qualitative validation made by 20 different users, we can state that our method achieves very good and representative summary results from the final user point of view. 

Additionally, and always considering that the ultimate goal of this project is to reactivate the memory pathways of MCI patients, it offers satisfactory results in terms of capturing the main events of the daily life of the wearable camera users.
A public-domain code developed for our visual summary methodology, is published in \footnote{\scriptsize

\url{https://imatge.upc.edu/web/publications/visual-summary-egocentric-photostreams-representative-keyframes-0}
}





\bibliographystyle{IEEEbib}
\bibliography{bibliography}

\end{document}